%% file: SemanticRLv2-ArXiv.tex
\documentclass[journal,commsoc,hidelinks]{IEEEtran}

\usepackage[T1]{fontenc} 
\usepackage{color}
\usepackage{multirow}
\usepackage{array}
\usepackage{amsmath}
\usepackage{amssymb}
\usepackage{amsthm}
\usepackage{mathrsfs}
\usepackage{url}
\usepackage{graphicx} 
\usepackage{stfloats}
\usepackage{enumitem}
\usepackage{algorithmic}
\usepackage{algorithm}
\usepackage{cite} 
\usepackage{svg}
\usepackage{bbm}

\ifCLASSOPTIONcompsoc
\usepackage[caption=false, font=normalsize, labelfont=sf, textfont=sf]{subfig}
\else
\usepackage[caption=false, font=footnotesize]{subfig}
\fi

\makeatletter
\newcommand{\removelatexerror}{\let\@latex@error\@gobble}
\makeatother

\graphicspath{} 

\def\eg{{e.g.}}
\def\ie{{i.e.}}
\def\etc{{etc}}
\def\ngram{{\textit{n}-gram~}}
\def\Figure{{Fig. }}
\def\etal{\emph{et al.}~}
\newcommand{\bh}[1]{\boldsymbol{\hat{#1}}}

\begin{document}
	\title{Reinforcement Learning-powered Semantic Communication via Semantic Similarity}
	\author{Kun~Lu,~Rongpeng~Li,~Xianfu~Chen,~Zhifeng~Zhao,~and~Honggang~Zhang
		\thanks{
			K. Lu, R. Li, and H. Zhang are with the College of Information
			Science and Electronic Engineering, Zhejiang University, Hangzhou 310027,
			China (e-mail: lukun199@zju.edu.cn; lirongpeng@zju.edu.cn;
			honggangzhang@zju.edu.cn).}
		\thanks{
			X. Chen is with the VTT Technical Research Centre of Finland, 90570 Oulu,
			Finland (e-mail: xianfu.chen@vtt.fi).}
		\thanks{
			Z. Zhao is with the Zhejiang Lab, Hangzhou 311121, China, and also
			with the College of Information Science and Electronic Engineering, Zhejiang
			University, Hangzhou 310027, China (e-mail: zhaozf@zhejianglab.com).}

	}

	\maketitle
	
	\begin{abstract}
		
		We introduce a new semantic communication mechanism - SemanticRL, whose key idea is to preserve the semantic information instead of strictly securing the bit-level precision.
		Unlike previous methods that mainly concentrate on the network or structure design, we revisit the learning process and point out the semantic blindness of commonly used objective functions. 
		To address this semantic gap, we introduce a schematic shift that learns from semantic similarity, instead of relying on conventional paired bit-level supervisions like cross entropy and bit error rate. However, developing such a semantic communication system is indeed a nontrivial task considering the non-differentiability of most semantic metrics as well as the instability from noisy channels. 
		To further resolve these issues, we put forward a self-critic reinforcement learning (RL) solution which allows an efficient and stable learning on any user-defined semantic measurement, and take a step further to simultaneously tackle the non-differentiable semantic channel optimization problem via self-critic stochastic iterative updating (SCSIU) training on the decoupled semantic transceiver. 
		We have firstly tested the proposed method in the challenging European-parliament dataset, which confirms the superiority of our method in revealing the semantic meanings, and better handling the semantic noise. Apart from the experimental results, we further provide an in-depth look at how the semantic model behaves, along with its superb generalization ability in real-life examples. An RL-based image transmission extension is also exemplified, so as to prove the generalization ability and motivate future discussion.
		
	\end{abstract}
	
	
	\IEEEpeerreviewmaketitle
	
	\section{Introduction}
	
	\IEEEPARstart{H}{ow} to transmit messages reliably has always been the topic where tremendous efforts are exerted. The key idea of this kind of communication is to secure a paired bit-level (or word/pixel-level) accuracy with well-designed system modules \cite{goldsmith2005wireless}. 
	Despite the technical advances, errors can occur when a clean channel is not guaranteed. 
	For example, when the transmitter (TX) sends ``I have just brought a yellow banana'', a receiver (RX) may get a message like ``I have brought a banana'', with the word ``just'' and ``yellow'' lost. Undesirable the above error is, one can still understand the general idea of what should have been expressed. However, unfortunately, not all mistakes share the same property - sentences like ``I have just brought a yellow banner'' and ``I have just borrowed a yellow banana'' both have a lower bit-level error rate, but they actually distort the true meanings severely, and should by no means be considered as a superb solution. 
	\label{Section1}
	
	We argue that a better solution is to introduce the concept and advantage of semantics. Suppose that we could find a precise semantic measurement, then it would be possible to distinguish one candidate with higher semantic score from those with higher bit-level accuracy but lower semantic similarity. 
	In this paper, we define the term \textit{semantic communication} as a system that tries to communicate the true meaning of a message, instead of being typically designed to ensure an exact recurrence for what has been transmitted. 
	
	In literature, the idea of effectively transmitting the key meanings can date back to Shannon and Weaver \cite{shannon1949mathematical}, known as one of the three levels of communication. 
	However, this very first endeavor does not gain much attention later on until in recent years with the rapid advance of machine learning and a growing interest in exploring semantic communication \cite{bao2011towards, strinati20216g, uysal2021semantic, qin2021semantic, shi2021new, lu2021rethinking}. 
	Among all related fields, the concept of \textit{semantic information} is first declared and discussed. Carnap \etal \cite{carnap1952outline} propose to calculate the semantic likelihood via logical probabilities and formulate a similar log-probability equation to measure the semantic information, being analogy with the classical information theory characterized by Shannon \cite{shannon1948mathematical}. This work is later refined by Floridi \cite{floridi2004outline} and D'Alfonso \cite{d2011quantifying} who further improve the way of measuring the \textit{truthlikeness} of a sentence. In \cite{bao2011towards}, Bao \etal investigate a general case of semantic information, and theorize certain key factors for semantic coding and semantic communication. Aside from quantitatively measuring the semantic information, there are also some works that explain from the perspective of biological system and physical world like \cite{johannsen2015semantic} and \cite{zhong2017theory}. Furthermore, theoretical analysis on the rate distortion and contextual inference are also investigated in recent literature \cite{liu2022task, seo2021semantics}.
	Meanwhile, researches on \textit{semantic similarity} have also gained an increasing attention.  
	Early approaches to measuring the semantic similarity are mainly based on the statistical analysis of contextual concepts \cite{jiang1997semantic, miller1991contextual}. More recently, the progresses on NLP (natural language processing) have brought more complicated and task-specific metrics, such as contextual word vector embedding \cite{mikolov2013efficient, pennington2014glove}, pre-trained language models \cite{devlin2018bert, brown2020language}, and non-parametric metrics \cite{papineni2002bleu, vedantam2015cider, zhu2016computing}. 
	We refer the readers to \cite{chandrasekaran2021evolution} for more details on semantic similarity in NLP. Nevertheless, it is worth mentioning that a general case of semantic similarity is not merely limited in sentences, but can be any media that express the semantics, or even for a task-execution purpose \cite{tung2021effective, maatouk2020age, ceran2021reinforcement}.

	In spite of the efforts on semantic information and semantic similarity, works on \textit{semantic communication} are still limited in the literature until in the latest few years. Early approaches mostly build the semantic communication process with probabilistic models and optimize in a collaborative or adversarial manner \cite{guler2018semantic, bao2011towards}. Farsad \etal \cite{farsad2018deep} then validate the feasibility of joint source-channel coding (JSCC) on text transmission by minimizing the word error rate. In DeepSC \cite{xie2021deep}, the authors further leverage an attentive language backbone to model the contextual semantic information and enable a varying-length semantic representation. Tung \etal \cite{tung2021effective}, on the other hand investigate the effectiveness problem from an RL-based collaborative perspective. Recently, a contextual inference scheme for semantic native communications is introduced in \cite{seo2021semantics}. Aside from the researches on communication schemes, advanced methods for improving the performance and flexibility, such as lite and distributed deployment \cite{xie2020lite}, hybrid automatic repeat request (HARQ) scheme \cite{jiang2021deep}, adaptive bit rate control \cite{zhou2022adaptive}, and semantic robustness \cite{hu2022robust} are also investigated. 
	Besides the aforementioned works that pursue a semantic recovery, another line of work aims to develop a semantics-aware communication for task-execution purpose. Specifically, efforts in this category concentrate more on the effectiveness (\ie, information bottleneck, transmission latency, throughout, overall performance) and are usually optimized to accomplish goal-oriented and collaborative tasks \cite{shao2021task, farshbafan2021common, yun2021attention, tung2021effective}.

	However, there are still several underlying problems that can not be properly solved with existing semantic communication approaches. First and foremost, most of the popular semantic communication schemes are concentrated on the model-level semantic extraction (\ie, with more advanced structures and transmission schemes), but their bit-level learning objective - a MSE or cross entropy (CE) loss inevitably introduces a semantic gap; thus they are more or less restricted within the scope of reliable transmission. Second, most works have now studied the differentiable objective optimization, but there are few  providing a universal framework that allows the optimization on \emph{any} non-differentiable one, which is however commonly seen in real wireless scenarios. Third, almost all existing works are built on the JSCC framework with a strong assumption of the differentiable channel model, or are designed to transmit only a few semantic tokens, which can hardly get satisfied or scaled in reality.

	In this paper, we jointly solve these critical yet challenging issues. Specifically to bridge the semantic gap, we introduce a systematic shift from pursing the bit-level accuracy to a closer semantic similarity, which takes a step further towards the goal of semantic transmission. 
	For the objective non-differentiability, we propose to find surrogate for the non-differentiable objective functions with semantic similarity-based reward signal from reinforcement learning. Considering the huge semantic space which is thousands-times larger than existing token-based RL communications schemes, a self-critic training method is further introduced to provide low-variance, stable, and efficient policy gradient for the whole system. 
	Furthermore to simultaneously tackle the wireless channel, we introduce a self-critic stochastic iterative updating (SCSIU) variant of SemanticRL by extending the proposed RL approach on both the TX and RX side, where the semantic encoding and decoding processes can be separately modeled with learnable Gaussian or softmax policies. This decoupled setting provides a complete, semantics-oriented, and non-differentiable surrogate for wireless semantic transmission with no extra parameters and a comparable performance as the commonly adopted JSCC scheme.

	We notice that although reinforcement learning (RL) has been widely adopted in networking tasks such as network slicing, resource \& routing scheduling, efficiency optimization \cite{li2018deep, xu2017deep, gadaleta2017d, luong2019applications}, only a few works managed to integrate it into the physical layer tasks like estimating the channel state \cite{oh2021channel}, decoding a few simple tokens \cite{carpi2019reinforcement}, or transmitting simple message representations for down-stream controlling tasks \cite{tung2021effective, yun2021attention}. Moreover, these methods either focus only on a specific topic, or are too limited to be extended in a large-scale case. A general-purpose and practical RL method is still absent in encoding-decoding scenarios in particular. 
	
	We summarize the contribution of this work as follows:
	\begin{itemize}
		\item Distinguished from the well-discussed problem of reliable transmission, we explore a novel schematic shift where the goal is not only to model the semantics, but also to communicate the semantics as accurate as possible.
		\item We put forward a practical solution by learning from non-differentiable semantic similarity to bridge the semantic gap. To enable a stable optimization, we further provide a new insight for conventional semantic communication schemes and introduce an RL-based optimization paradigm. Typically, a new self-critic policy gradient approach for large-scale and complex semantic transmission is introduced, which provides a precise and efficient gradient estimation whilst bringing no extra parameters. 
		\item We validate the effectiveness of the proposed semantic similarity-oriented JSCC solution (i.e., SemanticRL-JSCC) against conventional JSCC approaches. Experimental result show that our method better catches the high-order semantic similarities other than the bit-level accuracy, with a remarkable performance boost in prevalent semantic metrics. 
		\item We exemplify a complete large-scale wireless semantic communication paradigm by simultaneously integrating the proposed self-critc training into the decoupled transceiver. This variant tackles both the non-differentiable channel and objective problem with learnable policies in both the semantic encoder and decoder, and is trained end-to-end with self-critic stochastic iterative updating (SematicRL-SCSIU). We have investigated the stability and convergence performance, and validated its comparable capability. 
	\end{itemize}

	The remainder of this paper is organized as follows. Section II briefly introduces the commonly used communication model and technical baselines, followed by a detailed description of the proposed work in Section III. In Section IV, both the objective and subjective evaluations are given for comparison, where the robustness test, and extensive experiments on real-life examples are also carried out. The limitations and possible future work are identified in Section V. Finally, Section VI concludes this paper.

	\section{System Model and Problem Formulation}

	\subsection{System Model}

\begin{figure*}[t]
		\centering
		\includegraphics[width=18.1cm]{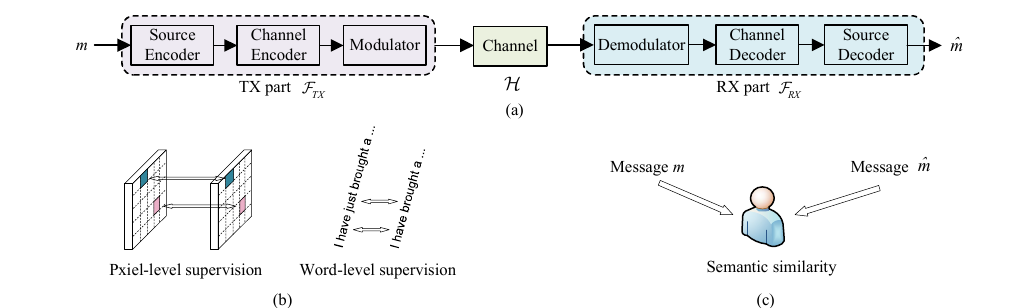}
		\caption{System model and comparisons of varied communication schemes. (a) Framework of a neural-network enabled communication system, where the TX part and RX part are modeled as abstract learnable modules. (b) The commonly used bit-level supervisions in existing joint source-channel coding schemes. (c) Our proposal. We develop a paradigm shift that communicates by learning from semantic similarity.}
		\vspace{-0.3cm}
		\label{Fig:COMM}
		\vspace{-0.1cm}
	\end{figure*}

	We consider a communication system with three major parts: TX side (parameterized by $\mathcal{F}_{TX}(\cdot)$), random channel ($\mathcal{H}(\cdot)$), and RX side ($\mathcal{F}_{RX}(\cdot)$), where TX and RX can contain multiple learnable modules and are tacitly assumed to be abstract in this paper. 
	Given a source message $\boldsymbol{m}$, we aim to recover it from the observed degraded \emph{d}-dimensional representation $\mathcal{F}^{\prime}_{TX}(\boldsymbol{m}) = \mathcal{H}(\mathcal{F}_{TX}(\boldsymbol{m})) \in \mathbb{R}^d$. The paradigm of such a transmission system is illustrated in \Figure \ref{Fig:COMM}(a). Also note that in this paper, we always parameterize the abstract encoder with $\phi$, and the abstract decoder with $\theta$. The detailed notations used in this paper are listed in Table \ref{tab:notations}.

	To enable a reliable transmission, mainstream schemes adopt pixel-level or word-level supervision like MSE and cross entropy, and trains the whole system in a deep learning-based JSCC manner (see \Figure \ref{Fig:COMM}(b)). To further facilitate the gradient back-propagation, the channel function $\mathcal{H}(\cdot)$ is usually modeled as additive or multiplicative ones \cite{o2017introduction, bourtsoulatze2019deep}. Thanks to the robustness and superiority 
	against separately designing of each of the module, JSCC methods are now gaining more attention, and have advanced the performance in a series of transmission tasks, 
	like those for images \cite{bourtsoulatze2019deep, shao2020bottlenet++}, texts \cite{farsad2018deep}, and speech signals \cite{weng2021semantic}.

\input{Tab_Notation}

	Different from the aforementioned popular schemes, a semantic communication system should concentrate more on the understanding and expression of semantics, instead of merely pursuing a simple bit-level recurrence. To formalize this task, we denote a semantic similarity metric by $\Theta$ and encourage the whole learning system $\{\mathcal{F}_{TX}, \mathcal{F}_{TX}\}$ to achieve higher semantic scores, as illustrated in \Figure \ref{Fig:COMM}(c). This is systematically different from conventional JSCC-based transmission pipelines or popular semantic communication schemes \cite{o2017introduction, bourtsoulatze2019deep, farsad2018deep, weng2021semantic} that interoperate the semantics from a model level but with bit-level supervisions. Our formulation, instead, evaluates how the semantics are represented and concentrates on the accuracy of semantic expression.

	As another key factor, the impact of random channel is a long-discussed yet troublesome problem. Although recent progresses on machine learning have provided plenty of state-of-the-art noise elimination schemes, like alternative training \cite{aoudia2019model}, GAN-based estimation \cite{ye2020deep}, and RL-based policy solution \cite{aoudia2018end}, these methods are usually computational intensive, or much sensitive to the parametric or environmental settings, which makes them hard to be integrated directly. Existing works on JSCC or semantic coding either make a strong hypothesis of an AWGN channel or directly sort to the off-the-shelf denoising techniques, with this fundamental problem only touched upon but not typically analyzed. In this paper, we take a step further by no longer assuming its differentiability, and aim to simultaneously tackle this problem with a unified learning framework.

	\subsection{Problem Formulation} 
	\vspace{-0.1cm}
	
	Hereinafter, we will mainly focus on the prevalent sentence transmission task unless specified. Typically, we denote such a message by $\boldsymbol{m}=\{w^{(1)},w^{(2)},...w^{(T)}\}$, where $T$ represents the sequence length, and each item $w^{(t)}$ is a single word coming from the dictionary $\mathcal{W}$. 
	For each word $w^{(t)}$ to decode, RX predicts the probabilistic distribution of choosing such a word on all candidates as a $V$-dimensional vector $[p(\hat{w}^{(t)}_1),p(\hat{w}^{(t)}_2),...p(\hat{w}^{(t)}_{V})]$, where $V$ denotes the range of all possible choices (\ie, the dimension of dictionary $\mathcal{W}$), based on the received signal $\mathcal{F}^\prime_{\textrm{TX}}(\boldsymbol{m})$. In the end, the decoded message is represented by $\boldsymbol{\hat{m}}=\{\hat{w}^{(1)},\hat{w}^{(2)},...\hat{w}^{(N)}\}$. Here we use $N$ to denote the length of decoded message $\hat{m}$, since it may be different from that of the sent message $m$ due to a noisy channel.
	
	To find the optimal transceiver (denoted by $(\cdot)^*$), the objective of our learning system is given by
	{\setlength\abovedisplayskip{0.05cm}
	\setlength\belowdisplayskip{0.05cm}
	\begin{equation}
		\langle \phi^*, \theta^* \rangle = \mathop{\textrm{argmax}}\limits_{\mathcal{F}_{TX}, \, \mathcal{F}_{RX}} \  \Theta(\boldsymbol{m}, \boldsymbol{\hat{m}})
		\label{Main}
	\end{equation}
	where we do not assume the differentiability of $\Theta(\cdot)$ like conventional JSCC methods do. This means that $\Theta(\cdot)$ can be \emph{any} reasonable semantic similarity metric. 
	
	\label{Bit-semantic}
	As a commonly-used metric, one can leverage the cosine distance of two message embeddings to quantize their similarity \cite{zhang2020context, meng2021magface, margaris2020adaptive}. 
	However, finding the most suitable embedding function is still an open question in this case. Although a learning-based embedding (\eg, that from a pre-trained network like BERT \cite{devlin2018bert}) is able to provide a task-specific representation, it is resource-consuming in both the training and inference stage, and is difficult to be generalized without a down-stream re-training. More significantly, using a black box as the guideline to train another black box, is problematic itself and not convincing enough (though we note that in practice one can also adopt it as $\Theta_{BERT}$ as we have no restriction on the choice of $\Theta$). Considering the above reasons, we choose the non-parametric NLP metrics like the commonly used BLEU \cite{papineni2002bleu} and CIDEr \cite{vedantam2015cider} score as an example when training the whole communication system. BLEU computes the \ngram similarity of two sentences based on the linguistic law that semantically consistent words usually come together in a given corpus, and are commonly adopted in exiting semantic communication works to evaluate the similarity score \cite{xie2021deep, jiang2021deep}, as in (\ref{NLP_BLEU}):
	{\setlength\abovedisplayskip{0.05cm}
	\setlength\belowdisplayskip{0.05cm}
	\begin{equation}
		\begin{aligned}
			&\Theta_{BLEU} = \textrm{exp}\left(1 - \frac{T_r}{\min(T_c,T_r)} + \sum^4_{n=1} \xi_n \log q_n \right)
			\label{NLP_BLEU}
		\end{aligned}
	\end{equation}
	where $q_n$ is the clipped \ngram score measuring the accuracy of \ngram expression in a decoded message $\hat{\boldsymbol{m}}$, and $\xi_n$ is the weighting parameter; $T_r$ and $T_c$ denote the length of ground truth reference and decoded candidate respectively. The first two terms in the power exponent are meant to introduce a punishment for short sentences. CIDEr score, take a step further to measure the cosine similarity of the ``term frequency inverse document frequency'' (TF-IDF) weights $g(\cdot)$ that measures both the accuracy and diversity of decoded \ngram phrases (where $(\circ)$ denotes the Hadamard product):
	{\setlength\abovedisplayskip{0.05cm}
	\setlength\belowdisplayskip{0.05cm}
	\begin{equation}
		\begin{aligned}
			&\Theta_{CIDEr} = \sum^{4}_{n=1} \xi_n \frac{g_n(\boldsymbol{m}) \circ g_n(\boldsymbol{\hat{m}})}
			{\Vert g_n(\boldsymbol{m})\Vert \cdot \Vert g_n(\boldsymbol{\hat{m}})\Vert }
			\label{NLP_CIDER}
		\end{aligned}
	\end{equation}
	
	In practice, BLEU and CIDEr are used to measure up to 4-gram similarity. A higher \ngram similarity measures more about the contextual information, while 1-gram score is a special case where the contextual information is totally ignored. It is expected that a semantic communication system concentrates more on the high order \ngram score, instead of the correctness of a single token. 
	
	Again, we highlight that through the introduced learning architecture (see \Figure \ref{Fig:FRAMEWORK-JSCC} and \ref{Fig:FRAMEWORK-COMP}), we pose no restriction on the choice of semantic similarity metric $\Theta$, which means SemanticRL accepts \emph{any} non-differentiable, task-specific, and reasonable similarity score, which includes but is not limited to the above-mentioned examples.

	Also note that as a special case of the proposed semantic communication system, we set $\Theta$ as the commonly-used (differentiable) negative cross entropy (CE) loss as the baseline when constructing numerical comparisons. CE formulates the sequence transmission as a classification task based on \textit{maximum likelihood} criteria, and is widely adopted in mainstream semantic communication works \cite{o2017introduction, xie2020lite, jiang2021deep, xie2021deep, farsad2018deep, zhou2021semantic}:
	\begin{equation}
		\Theta_{CE} = \log\, \prod_{t=1}^{T} p(w^{(t)}|\hat{w}^{(0)},...,\hat{w}^{(t-1)}; \phi, \theta)
		\label{CE_loss}
	\end{equation}

	In the following, we will elaborate on how to solve the problem formulated in (1) with either non-differentiable forms (\eg, (2), (3)) or simply differentiable form (4), where a non-differentiable channel is typically considered.

	\section{The SemanticRL Scheme in Non-Differentiable Environment}

	\subsection{The Markov Decision Process Framework for SemanticRL}
	In existing work as \cite{o2017introduction}, a neural network-enabled communication system learns from paired data and labels with cross entropy loss exerted on each of them. However, when trying to maximize the semantic similarity, there lines two critical challenges: 1) no labels are available during the recovery of each symbol until the whole sentence is derived on the RX side; and 2) most of the current semantic similarity metrics (like BLEU and CIDEr) are formulated on a prior statistical knowledge (frequency of occurrence, vector embedding, for instance) and thus they are usually non-differentiable. 
	
	To solve these issues, we take inspiration from another familiar game - the Go. A player will never know the final results until the end of a game, and to win or to loss is also hard to model using a regular loss function but in the form of \textit{reward} \cite{silver2016mastering}. Through the transformation from conventional classification task to a decision-making process, we make the first attempt to unfold an RL-powered semantic communication system, that optimizes directly from the semantic similarity, and concretely bridge the semantic gap beyond long-establish practice of bit-level supervision.

	First and foremost, we present the state-action definitions for the presented semantic transmission architecture:

	\begin{itemize}
		\item \textit{State}. We define the state as the recurrent state of decoder $s_{\textrm{de}}^{(t)}$ (see Section \ref{imp_details} later), combined with the history of actions taken so far. Therefore, we have $s^{(t)}=\{s_{\textrm{de}}^{(t)}, \hat{w}^{(0)},\hat{w}^{(1)}...,\hat{w}^{(t)}\}$. Once the next token $\hat{w}^{(t+1)}$ (the next word in a sentence transmission task for example) is generated, the transition between two adjacent states is deterministic.
		\item \textit{Policy}. In the sentence transmission task, we define the policy as which token to generate based on the current state. 
		The decision-making process terminates when the policy loop generates an unique token ``<EOS>''.
		\item \textit{Action}. The definition of action is quite intuitive - generating a new token from dictionary $\mathcal{W}$. However as determined by the semantic essence, the action space (and the resulting state space) is orders of magnitude bigger than conventional token transmission.
		\item \textit{Reward}. Since the proposed system is trained to directly optimize the semantic similarity, we use $\Theta(\boldsymbol{m},\bh{m}) \rightarrow \mathbb{R_+}$ to denote such a specific measurement.
	\end{itemize}

	\begin{figure*}[b]
		\centering
		\includegraphics[width=18.1cm]{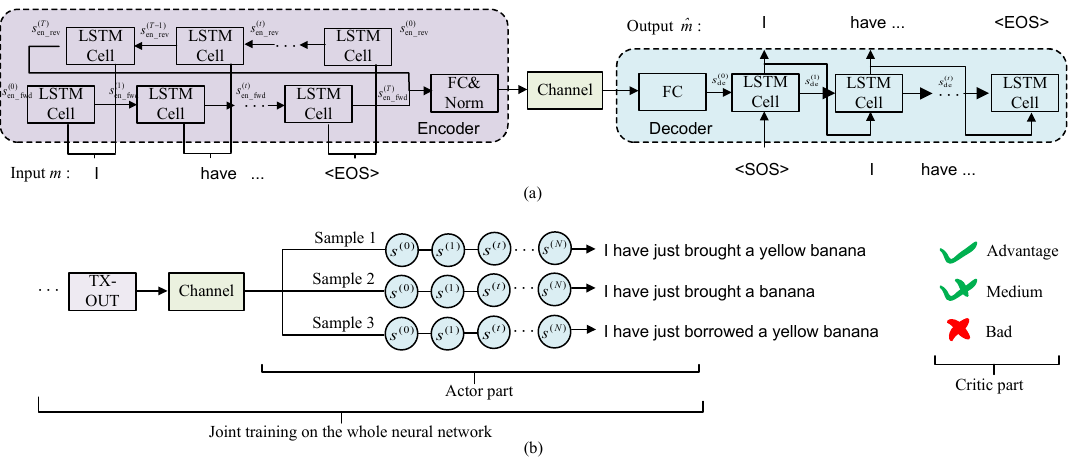}
		\vspace{-0.5cm}
		\caption{The detailed structure of SemanitcRL-JSCC. (a) We adopt a Bi-LSTM as the encoder and a LSTM as the decoder (shown in light purple and blue blocks respectively), where Norm represents the average power constraint. (b) Illustration of the self-critic optimization scheme. We formulate the decoding task as an episodic decision-making process (in this scheme, the policy is also conditioned on the encoder), and provide surrogate policy gradient for non-differentiable semantic similarity metrics. Through parallel sampling, the advantage semantic expression is rewarded and vice verse for the bad ones, which eventually leads to a precise and low-variance gradient estimation towards a policy improvement.}
		\label{Fig:FRAMEWORK-JSCC}
		\vspace{-0.3cm}
	\end{figure*}

	Without loss of generality, the objective function, or \emph{return} for the investigated system is formulated as: 
	\begin{equation}
		G^{(t)} = r^{(t+1)} + \gamma r^{(t+2)} + \gamma^2 r^{(t+3)} + ... = \sum_{k=0}^\infty \gamma^{k} r^{(t+k+1)}
		\label{def_Gt}
	\end{equation}
	where $\gamma$ is the discount factor. In the cases of sentences generation, we can always use $\gamma=1$ to simplify the discussion\cite{pitis2019rethinking}, as also adopted in many other sequence-generation tasks \cite{bahdanau2016actor, ren2017deep}. In literature, $G^{(t)}$ can also be defined conditioning on an agent's current state or together with its potential action, which is usually called the state-value function $V(s)=\mathbb{E} [G^{(t)}|s^{(t)}=s]$ and the state-action value function $Q(s,a)=\mathbb{E} [G^{(t)}|s^{(t)}=s,a^{(t)}=a]$, respectively.

	One unique property of the above strategy is that, the intermediate rewards can not be directly measured until at the end of a sentence. More specifically, the intermediate reward is always zero except for the last time step, which means the reward signal is \emph{sparse} across the decoding process:
	\begin{equation}
		r^{(t)} = \left\{
		\begin{aligned}
			0\quad\quad				\quad \textrm{if} \   t \neq N \\
			\Theta(\boldsymbol{m},\bh{m}) 	 	\quad \textrm{if} \   t = N
		\end{aligned}
		\right. 
		\label{reward}
	\end{equation}
	
	We point out that one solution is to implement Monte-Carlo rollouts for each time step as \cite{silver2016mastering} \& \cite{yu2017seqgan} do, or one can train another neural network to estimate the intermediate reward or the reshaped reward for an incomplete sequence \cite{konda2000actor}, which is known as an \textit{actor-critic} way. However, the drawbacks of these methods are also evident: the former one is much time-consuming, while the other one is resource-consuming; and both face the risk of divergence in a huge action and state space which in our scenario is more than $10^4$. 
	
	\vspace{-0.1cm}
	\subsection{Optimization in a Self-critic Manner}

	\subsubsection{SemanticRL-JSCC}

	To give a more general derivation, we first assume that the intermediate reward $r^{(t)}$ is non-zero, and the discount factor $\gamma$ is set to $1$ as mentioned before. Denoted by $\phi$ and $\theta$ the parameters of the encoder and decoder respectively, the objective function for such a JSCC-based semantic transmission process from $s^{(0)}$ is given by maximizing
	
	\begin{equation}
		\begin{aligned}
			J(\phi; \theta) &= \mathop{\mathbb{E}}\limits_{\hat{w}^{(1)},...,\hat{w}^{(N)}} \left[ G^{(0)}|s^{(0)} \right] \\
			&= \mathop{\mathbb{E}}_{\hat{w}^{(1)},...,\hat{w}^{(N)}} \left[ \sum_{t=1}^{N}  r^{(t)} \right]
		\end{aligned}
		\label{objectivef}
	\end{equation}
	where $\hat{w}^{(1)},...,\hat{w}^{(N)}$ is one complete trajectory, \ie, one complete sentence generated from the initial state $s^{(0)}$. The objective function $J(\theta)$ can be interpreted as the expected total reward starting from the initial state. Note that in a JSCC-based learning procedure, $\phi$ and $\theta$ are jointly updated towards a higher semantic score. Thus, the optimal parameter of the semantic communication system can be found by $\langle \phi^*, \theta^* \rangle = \mathop{\textrm{argmax}} J(\phi; \theta)$.
	
	To enable such a transmission system, we give the optimization direction, or the \emph{gradient} of (\ref{objectivef}) by the following theorem:

	\noindent
	\textbf{Theorem 1.} \textit{(Semantic Policy Gradient) Denote by $\pi_{\phi, \theta}$ the policy that generates a complete sequence $\hat{w}^{(1)},...,\hat{w}^{(N)}$ each with probability $\pi_{\phi, \theta}(\hat{w}^{(t)}|s^{(t)})$, the gradient of $J(\phi; \theta)$ can be approximated via the following Monte-Carlo estimation:}
	\begin{equation}
	\nabla J(\phi; \theta) \approx \sum_{t=1}^{N} \nabla_{\phi, \theta} \log \pi_{\phi, \theta}(\hat{w}^{(t)}|s^{(t)}) \Theta(\boldsymbol{m},\bh{m})
	\label{probab_traj}
	\end{equation}

	\emph{Proof:} The proof is provided in Appendix A.

	\noindent
	\textbf{Corollary 1.} \textit{For continuous control tasks (\ie, an agent is asked to act in a continuous domain), Theorem 1 also holds true with a continuous policy $\pi: \mathcal{S} \times \mathcal{A} \rightarrow \mathbb{R}$.}

	Although Monte-Carlo rollout provides an unbiased estimation for the expected return, sampling from a large space still introduces a high variance for (\ref{probab_traj}). As a commonly-used technique, the above equation can be further simplified and improved by adding a baseline term $b$ for the accumulated reward \cite{williams1992simple}.

	In the \textit{actor-critic} algorithm, the baseline term is modeled as the state-value function $V(s^{(t)})$. However, such a training scheme will inevitably need to train another value network so as to provide a precise estimation for $V(s^{(t)})$. 
	Similarly, some other works treat the estimation of $b$ as another prediction task \cite{ranzato2015sequence}, but the optimization process still involves certain extra parameters, making the whole framework sensitive to the correctness of bias estimation and more difficult to train.
	
	In this paper, we resort to a simpler and quicker solution by using the mean \textit{return} from a group of parallel samples, \ie, $\{\bh{m}_1, \bh{m}_2.., \bh{m}_{M-1}\}$ as the baseline term, where $M-1$ is the number of selected samples. This trick is usually called ``self-critic'' in literature \cite{rennie2017self, luo2020better}, since neither does it need to carry out Monte-Carlo rollouts for each time step, nor does it require another neural network to estimate the baseline term, but only calls for more parallel samples. 
	
	\label{sec:selfcrit-decoder}
	To generate a collection of sample, we no longer model the choice of next token (\ie, an \emph{action}) as a static distribution followed by argmax operation (as commonly used in existing semantic communication works). Instead, we model the output likelihood as a probabilistic \emph{multinomial distribution} (denoted by $\mathcal{MD}$), as illustrated in (\ref{sample_decoder}). Note that sampling from this distribution ensures that those with maximum likelihood is more frequently selected, and simultaneously allows certain exploration during the decision-making process. As such, we can always obtain the aforementioned \emph{M}-1 samples to provide a proper, and precise estimation for the baseline term. 
	
	\begin{equation}
	\pi^{(t)}_{\phi, \theta} := \textrm{Sample}\left(\mathcal{MD}\left(\left[p(\hat{w}_1^{(t)}), p(\hat{w}_2^{(t)})...p(\hat{w}_V^{(t)})\right]^T \right)\right) 
	\label{sample_decoder}
	\end{equation}
	where we here use the superscript $t$ explicitly to highlight that the sampling process happens at all time steps. However note that the policy network is not updated until at the end of a complete transmission.
	
	Finally, we calculate the gradient of $J(\phi; \theta)$ as follows:
	
	\begin{equation}
		\begin{aligned}
		\nabla J(\phi; \theta) \approx \frac{1}{M}  \sum_{i=1}^M \left[ \sum_{t=1}^{N}  \right. \nabla_{\phi, \theta} &\log \pi_{\phi, \theta, i} (\hat{w}^{(t)}|s^{(t)}) \cdot
		\\
		&\left. \left( \Theta_{i} - \mathop{\textrm{avg}}\limits_{k\sim M;\ k\neq i} (\Theta_{k}) \right) \right]
		\label{self-critic}
		\end{aligned}
	\end{equation}
	where $\textrm{avg}(\cdot)$ calculates the mean value, and the subscript $i$ in $\pi_{\phi, \theta, i}$ implies the $i$-th parallel policy. We use $\Theta_{k}$ to denote the \textit{return} of a parallel sample $\bh{m}_k$. 

	The right most bracket in (\ref{self-critic}) can be viewed as an advantage where expressions resulting in a higher semantic similarity are given a positive reward, while those with a low semantic similarity are therefore punished (also see \Figure \ref{Fig:FRAMEWORK-JSCC}(b)). In a practical communication system, (\ref{self-critic}) can be extremely important as it enables a fast and stable training at the cost of nearly no extra computations. 
	
	Step by step, we summarize the critical components of SemanticRL-JSCC in Algorithm 1.
	
	\input{Alog_SemanticRL-JSCC}

	\begin{figure*}[b]
		\centering
		\includegraphics[width=18.1cm]{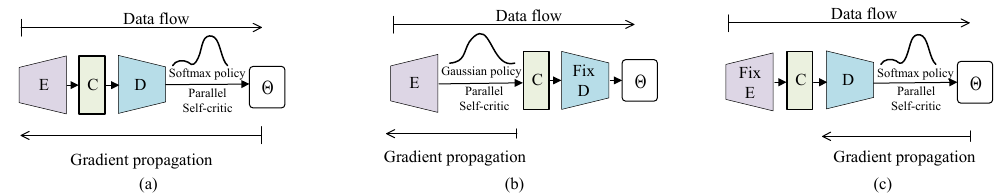}
		\caption{Comparisons and details of the proposed two variants, where E, C, D denote the encoder, channel, decoder respectively. (a) In the JSCC variant, the policy is dependent on both the encoder and decoder; thus, the system is correspondingly jointly optimized  with a RX-side softmax policy. In (b) and (c) we illustrate the encoder training and decoder training scheme in SemanticRL-SCSIU respectively, where these two parts are decoupled and independently optimized with different policies (Gaussian and softmax) but toward the same semantic goal. The presented scheme introduces no extra parameter as conventional approaches, but learns directly from semantic similarity and jointly copes with the objective-level and channel-level non-differentiability.}
		\label{Fig:FRAMEWORK-COMP}
	\end{figure*}

	\subsubsection{SemanticRL-SCSIU}
	\label{section:SCSIU}
	In the above discussion, we introduce an RL-powered semantic communication system that directly optimizes the similarity metrics. However in a JSCC manner, the channel function is still assumed to be differentiable and does not always hold true in real scenarios. To simultaneously deal with the non-differentiability of random channels, we can further introduce RL training into the TX side and turn the whole learning system into a collaborative semantic transceiver. Prior to our work, the idea of training an RL-based encoder was once investigated by Aoudia \etal \cite{aoudia2018end} and Tung \etal \cite{tung2021effective}, but these works are only concentrated on transmitting a few symbols and could hardly get scaled to a large deployment due to the difficulty of training another value network or providing precise reward estimations, especially in semantic transmissions where the action space and states are thousands-times larger.
	
	To handle the large-scale extensibility issues, we introduce a self-critic training scheme into the training of both the TX and RX side, and iteratively update these components. In this setting, the semantic encoder and decoder are now decoupled as two independent agents but optimized with a shared objective - to maximize $\Theta(\boldsymbol{m}, \bh{m})$.
	
	Specifically, and different from the JSCC variant introduced in Section \ref{sec:selfcrit-decoder}, we freeze $\phi$ and view the output of $\mathcal{F}_{TX}$ as a new virtual dataset when optimizing the decoder. In the decoder training stage, the policy and objective function is now only parameterized on $\theta$. That is
	
	\begin{equation}
		\pi^{(t)}_{\theta} := \textrm{Sample}\left(\mathcal{MD}\left(\left[p(\hat{w}_1^{(t)}), p(\hat{w}_2^{(t)})...p(\hat{w}_V^{(t)})\right]^T \right)\right) 
		\label{sample_decoder-SCSIU}
	\end{equation}

	On the other hand in the encoder training, the action space is now a continuous one, and has only one time step (a special case of the above derivation). From Corollary 1, we learn that the same self-critic scheme can be adopted as well except that we now model the policy with a continuous Gaussian distribution (denoted by $\mathcal{N}$). Therefore, we keep following the same routine introduced in Section \ref{sec:selfcrit-decoder} but provide a self-critic Gaussian policy gradient when optimizing the semantic encoder. In this setting, the output embedding of $\mathcal{F}_{TX}$ is then converted to the \emph{mean} value $\boldsymbol{\mu}^{d\times 1}$, and we set the covariance matrix $\boldsymbol{\Sigma}^{d\times d}$ as ${(\sigma\boldsymbol{I})}^2$ to encourage certain exploration (typically we set $\sigma$ as 0.1):
	
	\begin{equation}
		\pi_{\phi} := \textrm{Sample}\left(\mathcal{N}\left(\boldsymbol{\mu}=\mathcal{F}_{TX}\left(\boldsymbol{m}\right), \  \boldsymbol{\Sigma}= {(\sigma\boldsymbol{I})}^2 \right)\right)
		\label{sample_encoder}
	\end{equation}

	To obtain the expected reward (or semantic similarity score in our case), a Monte-Carlo simulation is implemented on the RX side. Note that when feeding forward through the decoder, we freeze its parameters and execute greedy decoding (via argmax) which makes the untrained part deterministic. Formally, we present the objective for both TX and RX part in (\ref{SeamnticRL-SCSIU})
	\begin{subequations}
	\begin{align}
	&\phi^* = \mathop{\textrm{argmax}}\limits_{\mathcal{F}_{TX}} \Theta(\boldsymbol{m}, \underbrace{\mathcal{F}_{RX}(\mathcal{H}}_{\textrm{no grad}}(\mathcal{F}_{TX}(\boldsymbol{m})))) \\ 
	&\theta^* = \mathop{\textrm{argmax}}\limits_{\mathcal{F}_{RX}} \Theta(\boldsymbol{m}, \mathcal{F}_{RX}(\underbrace{\mathcal{H}(\mathcal{F}_{TX}(\boldsymbol{m})}_{\textrm{no grad}}))) 
	\end{align}
	\label{SeamnticRL-SCSIU}
	\end{subequations}

	Note that through self-critic training, we provide a parallel and precise gradient surrogate, where we neither needs to train $any$ other module to facilitate the learning process as in \cite{tung2021effective}, nor suffer from a poor estimation of the expected \emph{return} as in \cite{aoudia2018end}. The detailed training scheme and gradient flow are illustrated in \Figure \ref{Fig:FRAMEWORK-COMP}.

	Finally, the gradient of log-probabilities for (\ref{sample_decoder-SCSIU}) and (\ref{sample_encoder}) can be calculated from the following theorem: 
	
	\noindent
	\textbf{Theorem 2.} \emph{(Policy Gradient Propagation) Let $\mathbbm{1}^{1\times V}$ be the indicator vector where only the sampled token's position are 1 otherwise 0,  $\mathcal{F}^{(t)}_{RX}(\cdot)\in\mathbb{R}^{V\times 1}$ be the output of RX at time step $t$, and $\widetilde{\mathcal{F}_{TX}}(\boldsymbol{m})$ the sampled message embedding, the gradient propagation of SemanticRL-SCSIU is given by}
	\begin{subequations}
	\label{Grad-SCSIU}
	\begin{align}
	&\nabla_\theta \log(\pi_{\theta}(\hat{w}^{(t)}|s^{(t)})) = \notag \\ 
	&\quad \quad \quad \left[ \mathbbm{1}^{1\times V} - \left[p(\hat{w}_1^{(t)}), p(\hat{w}_2^{(t)})...p(\hat{w}_V^{(t)})\right] \right] \left[\nabla_\theta \mathcal{F}^{(t)}_{RX}\right]  \\
	&\nabla_\phi \log(\pi_{\phi}) =  \left[\widetilde{\mathcal{F}_{TX}}(\boldsymbol{m})-\mathcal{F}_{TX}(\boldsymbol{m})\right]^T \boldsymbol{\Sigma}^{-1}
	\left[\nabla_{\phi} \mathcal{F}_{TX}(\boldsymbol{m})\right]
	\end{align}
	\end{subequations}
	
	\emph{Proof:} The proof is provided in Appendix B.
	
	The detailed procedures for training SemanticRL-SCSIU can be found in Algorithm 2.

\input{Alog_SemanticRL-SCSIU}

	\section{Experiments}

	\subsection{Implementation Details}
	\label{imp_details}
	\subsubsection{Neural Network Settings}
	To demonstrate the effectiveness of the proposed RL-based scheme, we refer to the work of Farsad \etal \cite{farsad2018deep} as the baseline model, where a Bi-directional LSTM is used as the encoder, and another LSTM as the decoder. On the RX side, the receiver initializes its hidden state from the received message $\mathcal{F}^\prime_{\textrm{TX}}(\boldsymbol{m})$, and then sequentially generates a token by conditioning on the previously generated one. The general framework is described in \Figure \ref{Fig:FRAMEWORK-JSCC}, where we denote the hidden state of forward and backward LSTM by $s_{\textrm{en\_fwd}}^{(t)}$ and $s_{\textrm{en\_rev}}^{(t)}$ in the encoder, and $s_{\textrm{de}}^{(t)}$ on the RX side.

	In practice, we will always model the message embedding $\mathcal{F}^\prime_{\textrm{TX}}(\boldsymbol{m})$ in complex representations \ie, $\left[ \mathcal{F}^\prime_{\textrm{TX}}(\boldsymbol{m})\left[0:d/2\right] + j \cdot \mathcal{F}^\prime_{\textrm{TX}}(\boldsymbol{m})\left[d/2:d\right]\right]^T$, and pose an average power constraint at the end of a transmitter. 
	As to the channel settings, we have considered both the commonly used AWGN channel and phase invariant fading channel, as in \cite{kurka2020deep}. 
	Note that the introduced training schemes (\ie, SemanticRL-JSSS and SemanticRL-SCSIU) will not introduce \emph{any new parameter} for the baseline CE model; thus the structure and parameters of these two models are identically the same except for their training methods. The detailed network structure is provided in Table \ref{tab:structure}.
	
	\begin{table}[b]
		\centering
		\caption{Network Structure of SemanitcRL. We omit non-learnable parts like channel and normalization for tidiness.}
		\label{tab:structure}
		
		\def\arraystretch{1.0}
		
		\begin{tabular}{ccc}	
			\hline
			Layer&Dimension&Details\\
			\hline
			\hline
			~&Encoder&~	\\
			\hline
			Shared Embedding&24064 $\times$ 128 & Dictionary size\\
			BiLSTM (2 LSTM Cells)&128 $\times$ 128 (2 layer) & Tanh activation\\
			Linear&256 $\times$ 256 & ReLU activation\\
			Linear&256 $\times$ 256 & Gaussian policy \\
			\hline 
			~ & Random Channel & ~ \\
			\hline
			~&Decoder&~	\\
			\hline
			Linear & 256 $\times$ 256 & Initialize LSTM Cell \\
			Shared Embedding&24064 $\times$ 128 & Shared knowledge \\
			LSTM Cell&128 $\times$ 128 (2 layer) & Tanh Activation\\
			Linear&128 $\times$ 128 & Dropout=0.5\\
			Linear&128 $\times$ 24064 & Dictionary size \\
			Softmax & None & Softmax policy \\
			\hline
			
		\end{tabular}
	\end{table}

	We note that a simple scheme like LSTM sufficiently demonstrates the benefits of introducing SemanticRL. However, one can also adopt more complicated language models like Transformer \cite{vaswani2017attention}. Similarly, we need to clarify that in this paper, we are more concentrated on the introduced semantic similarity-based communication schemes along with the corresponding learning methods, but not the detailed choice of a specific similarity score $\Theta$, which we suggest can be designed and adjusted for different down-stream tasks (see Section \ref{sec:choiceofTheta}). From the technical and deployment perspectives, we tend to use small models to keep the communication system compact and efficient; and the proposed approach is both model-agnostic and objective-agnostic. 
	
	\subsubsection{Training}
	\label{Training}
	Although (\ref{self-critic}) theorizes the way to a stable training in our semantic communication system, the huge action and state space (typically the action space is more than $10^4$ compared with concurrent RL-based works that concentrate on dozens of symbols) 
	remains one critical challenge at the very beginning of the training process. To facilitate the training process, we follow a widely accepted protocol 
	by first pre-training a model with regular loss functions (\eg, MSE and cross-entropy loss), and then use reinforcement learning to proceed the rest training \cite{rennie2017self, yun2017action}. For fairness, we seek to initialize the parameters from a partially-trained CE model (for the first 87 epochs) so that we can better visualize the difference between conventional approaches and the newly introduced one. The learning rate of this CE model is set to $1e-3$ initially and dropped by half at epoch 20. The resulting parameters are used to initialize both the baseline model and our SemanticRL. In the second stage, we train the baseline model and SemanticRL-JSCC separately with a learning rate of $1e-4$ till epoch 160, where the learning rate is again reduced by half. Training process terminates at epoch 200 for both examined approaches. For SemanitcRL-SCSIU, we keep the learning rate as 1e-4 and train it till epoch 300, which is about two times longer considering that only half on the parameters are updated in a minibatch. As to the semantic similarity metric $\Theta$, we choose CIDEr-D by default as it proves to be effective in many other sentence-matching tasks \cite{liu2017improved, rennie2017self}, when not specified. The batch size is set to 64 as a constant for all implementations. 
	
	We provide a PyTorch implementation to verify the results reported in this paper\footnote{\url{https://github.com/lukun199/SemanticRL}}.

	\begin{figure*}[b!]
		\centering
		\includegraphics[width=18.1cm]{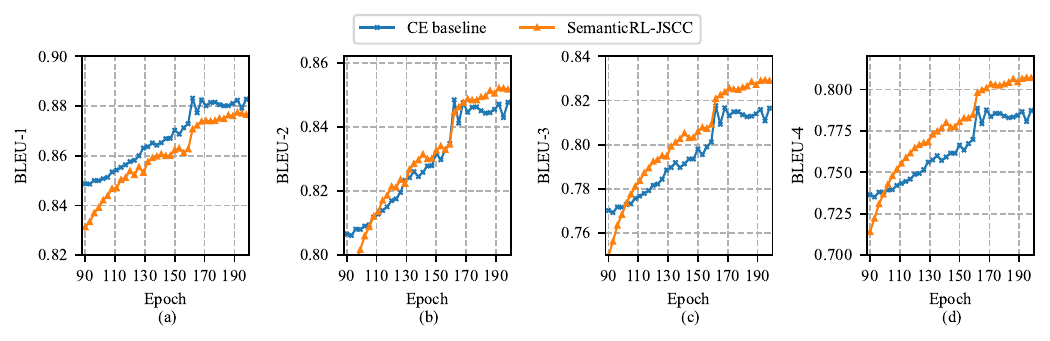}
		\caption{Semantic communication performance on BLEU metric. Scores on BLEU-1 to BLEU-4 are respectively presented in (a)-(d). As we put more emphasis on contextual semantics (high order \ngram similarity), the advantage of SemanticRL gets increasingly more evident.}
		\label{Fig:BLEU}
	\end{figure*}
	
	\begin{figure*}[t!]
		\centering
		\includegraphics[width=18.1cm]{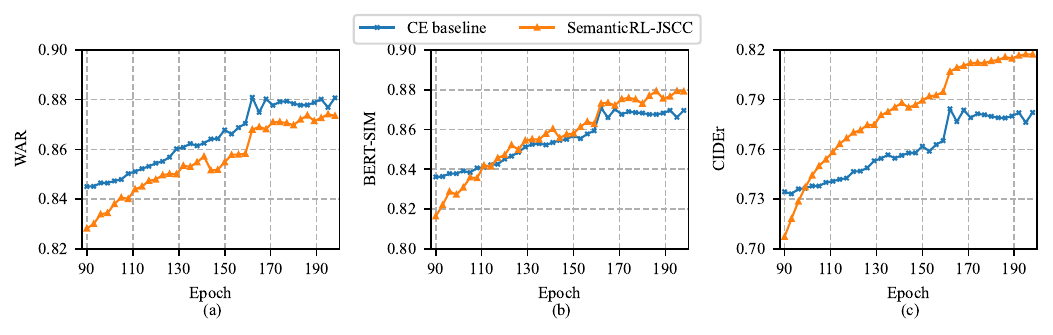}
		\caption{Quantitative comparisons on word accuracy rate (WAR), BERT similarity (BERT-SIM), and CIDEr score. SemanticRL learns stably and catches more semantic infraction at the expense of certain bit level accuracy. The performance drop on early epochs is caused by the shift on learning objective.}
		\label{Fig:MROE_SCORE}
	\end{figure*}

	\begin{figure}[t]
		\centering
		\includegraphics[width=8.8cm]{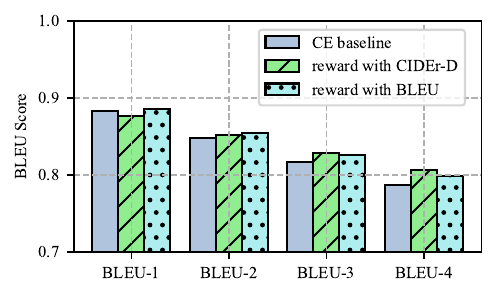}
		\caption{Comparisons on the choice of semantic similarity metrics. The optimized scores correspondingly witness a performance boost; thus enabling a task-specific and customized optimization.}
		\label{Fig:BLEU_REWARD}
	\end{figure}
	
	\subsection{Datasets and Metrics}
	
	We use the standard dataset of European Parliament \cite{koehn2005europarl} as our baseline model and some other previous works have done \cite{xie2021deep, farsad2018deep}. This dataset contains more than 2 million sentences with a huge dictionary of around 54 million tokens. We perform a simple pre-processing for this dataset, where all words are turned into the lowercase and the punctuation marks are removed. Considering the long-time dependency of our language model, as reported in \cite{farsad2018deep}, sentences with length from 3 to 20 are collected; meanwhile words appearing less that five times are replaced with the ``<UNK>'' token. 
	In our experiments, the processed dataset contains a total of 886,986 sentences and 24,064 vocabularies. It is then divided into training and testing split with a ratio of 4:1.

	Being in line with existing papers on semantic communication, we also use BLEU score and the cosine similarity from BERT embeddings (BERT-SIM) to measure the semantic similarity. Moreover, we report the results on CIDEr and word accuracy rate (WAR, implemented as the ratio of correct word pairs to the maximum sentence length) as complementary metrics. Note BERT model is task-specific, and may cause certain instability and generalization problem if not fine-tuned with a down-stream, task specific manner. Here we adopt a popular open-source implementation\footnote{\url{https://github.com/UKPLab/sentence-transformers}} that is trained on a paraphrase task to calculate the similarity scores, which is shown to be more sensitive and precise than the original BERT model or general purpose BERTs.
	
	Note that for fairness, we will always report the results of SemanticRL-JSCC unless specified, as almost all the semantic communication approaches, including our baseline model are trained with this routine.

	\subsection{Numerical Results}
	
	\subsubsection{On Semantic Similarity}

	We report the runtime BLEU score of SemanticRL-JSCC in \Figure \ref{Fig:BLEU}, which is compared with the CE-based baseline model. BLEU score (given in (\ref{NLP_BLEU})) counts the accuracy of a \ngram phrase between the reference and candidate query sentences, where the results from 1-gram to 4-gram are known as BLEU-1,2,3, and BLEU-4, respectively. First, we observe the proposed method, as notated in orange lines, learns stably as the training proceeds, which firmly demonstrates the feasibility of an RL-based non-differentiable optimization. Specifically, SemanticRL gradually outperforms the CE baseline model in BLEU-2/3 and BLEU-4, with a comparable result in BLEU-1. These results are rather interesting, given that a longer phrase contains more about the context, while a short phase considers more on the word-level accuracy, which on the other hand provides a strong evidence for the semantics-preserving capability of the proposed approach. On the contrary, BLEU-1 score, as can be found in \Figure \ref{Fig:BLEU}(a) is almost the same as the word accuracy rate (WAR) in \Figure \ref{Fig:MROE_SCORE}(a). Although the baseline model (with CE loss on each word pair) produces a higher BLEU-1 score, long-term semantic consistencies are sacrificed to some extent, as our discussions in Section \ref{Bit-semantic} have revealed. Note that we do not emphasis the superiority on all the n-gram scores, but a schematics shift and improvement from low \ngram to high-order semantic similarity, which better reveals the contextual semantic meanings.

	\begin{figure*}[t]
		\centering
		  \subfloat[]{%
			\includegraphics[width=18.1cm]{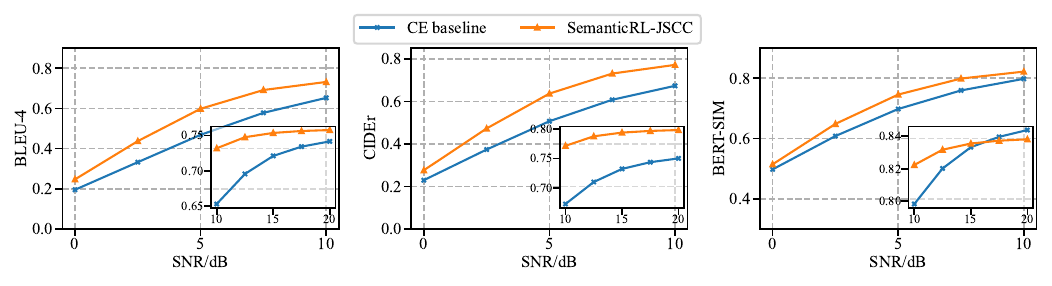}}
		\hfill
		\subfloat[]{%
			\includegraphics[width=18.1cm]{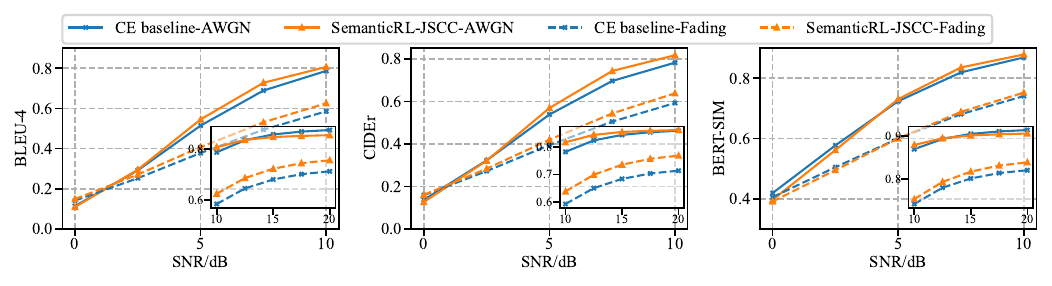}}
		\caption{Evaluations on the generalization ability and robustness. (a) Quantitative results on a varying AWGN channel where there is no training and inference gap. The main body reports the results from 0 to 10 dB, while the lower right windows exhibit the results from 10 dB to 20 dB. (b) Generalization on a fixed 10 dB training and varying-SNR inference in both AWGN and phase invariant fading (FIF) channels.}
		\label{Fig:VARY_SNR}
	\end{figure*}	
	
	\begin{figure*}[t!]
		\includegraphics[width=18.1cm]{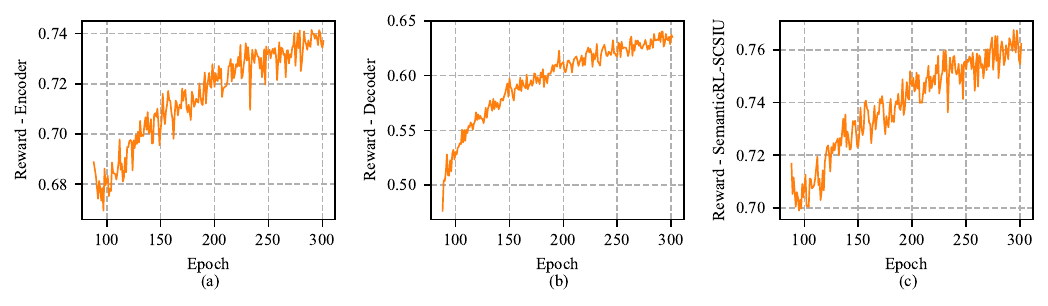}
		\vspace{-0.4cm}
		\caption{Convergence performance of SemanticRL-SCSIU (self-critic stochastic iterative updating). From (a) to (b), we provide the runtime reward (CIDEr-D score) of the encoder, decoder, and whole network respectively. The model is trained following Algorithm 2.}
		\label{fig:scsiu}
	\end{figure*}

	Besides BLEU score, we next investigate the inception scores derived from pretrained BERT extractor in \Figure \ref{Fig:MROE_SCORE}(b). The scores of SemanticRL take a similar pattern that increase stably and outperforms the baseline model. It is worth mentioning that although we do not explicitly use $\Theta_{BERT}$ as the reward function, BERT-SIM also supports the superiority of designing such a semantics-targeted communication system. This is because conventional approaches, in spite of the help from an advanced language model (or backbone network if deployed in transmitting other media like images and speech signals) that models the contextual information, is actually misguided by a non-semantic learning objective that ultimately hampers a semantic level communication.

	Another interesting feature of the proposed method is that the score on the corresponding training protocol is significantly higher (see \Figure \ref{Fig:MROE_SCORE}(c) for example), which is not hard to understand since the model is directly optimized with such a criterion. We argue that it may be hard in practice to find a model that fully outperforms the others in all the quantitative metrics, as these metrics each concentrates on different aspects of a given sentence pair. Nevertheless, one can freely choose any target reward to optimize for a given task. Accordingly, we provide an example in which the model is optimized with a combination of BLEU-1 and BLEU-4 score in \Figure \ref{Fig:BLEU_REWARD} to better demonstrate this point. The model trained with a specific similarity criterion behaves favorably toward the corresponding metric, which usually brings a notable performance boost. 
	
	In real applications, the choice of a semantic similarity metric $\Theta$ is largely determined by the specific task. We highlight that SemanticRL provides the first universal approach and systematic shift towards semantic transmission, but it is impossible to give each possible metric a try. Meanwhile, the design of semantic metrics, itself, is an important task in both semantic information theory and quantitative evaluations; and, as we show in this paper can be directly optimized to facilitate such a communication system. 
	
	\label{sec:choiceofTheta}

	\subsubsection{On the Impact of a Noisy Channel}

	We first investigate the robustness of the proposed model under a varying AWGN channel, where the SNR levels randomly fluctuate from 0 to 20 dB. \Figure \ref{Fig:VARY_SNR}(a) presents the BLEU-4 score, CIDEr, and BERT similarity scores for both our JSCC variant and CE baseline, where the results in 10 to 20 dB are cropped into a small window at the bottom right part of each sub-figure (in most cases we are more concentrated on low SNRs). It can be found that a) the proposed SemanticRL approach consistently outperforms the CE baseline in revealing the semantics; b) in near-boundary SNR levels (\ie, 0 dB and 20 dB), the performance gaps are generally lower the those around 10 dB, which indicates that the proposed model learns to develop a balanced strategy for both low- and high-SNR situations, instead of merely memorizing one specific SNR region and forming a biased transmission. c) SemanticRL is shown to better handle low-SNR situations (typically, 2.5 dB to 7.5 dB) than the baseline model, which hints that a similarity-targeted reinforcement learning scheme has the potential to better combat semantic noise and are probably more robust.

	While the above experiment provides certain evidence for the semantic robustness, we carry two more experiments to further examine the extent to which the introduced approach degrades under different testing scenarios. First, we formulate a reference benchmark where both models are trained under AGWN channel with a fixed SNR level of 10 dB. Then, we compare their performances by a) generalization on other unseen SNR levels; and b) training on more complex phase invariant fading (FIF) channels on top of setting a). In \Figure \ref{Fig:VARY_SNR}(b), we illustrate these experimental results. It can be observed that in both AWGN and FIF situations, SemanticRL exhibits certain generalization capacity for even unseen SNR levels. Specifically from 5 dB to 12.5 dB, it retains an obvious advantage over the baseline model with almost negligible degradation on the superiority. Furthermore, we have also found from the dashed lines that the proposed model presents a more evident advantage when tested under complex channel states, while the CE baseline suffers from a more severe semantic degradation. This observation further verifies the semantic robustness of the introduced framework, and its superiority on exploring semantics-invariant representations.

	\subsubsection{Joint Training vs. Iterative Updating}

	In all the above experiments, we present the comparisons with CE baseline in a JSCC manner. 
	Recall that the distortion introduced by a random channel $\mathcal{H}(\cdot)$ is usually unpredictable and can not be modeled with a closed-form function, JSCC training is actually a considerably ideal and strong assumption adopted in popular semantic communication schemes. In the SCSIU (self-critic stochastic iterative updating) variant of SemanticRL, we render it possible to eschew the non-differentiability by extending RL training in both semantic coding and semantic decoding, and modeling the communication process through the interactions with a noisy environment. Figure \ref{fig:scsiu} shows the runtime \emph{return} of this approach. It is vividly shown that both the TX and RX side are able to learn stably and tend to converge as the training proceeds.

	\begin{table}[t!]
		\centering
		\caption{Comparisons on the variants of SemanticRL. SCSIU-PT is the SCSIU training on top of the CE pretrained parameters, while SCSIU-FT implies fine tuning on top of SemanitcRL-JSCC. B@ represents the BLEU score here.}
		\label{tab:SCSIU_res}
		
		\def\arraystretch{1.0} 
		
		\begin{tabular}{p{1.4cm}<{\centering}p{0.7cm}<{\centering}p{0.55cm}<{\centering}p{0.55cm}<{\centering}p{0.55cm}<{\centering}p{0.55cm}<{\centering}p{1.4cm}<{\centering}}			
			\hline
			Variant&B@1&B@2&B@3&B@4&CIDEr&BERT-SIM\\
			\hline
			JSCC&0.877&0.852&0.829&0.807&0.817&0.879\\
			SCSIU-PT&0.849 &0.812 &0.780 &0.749 &0.747 &0.840 \\
			SCSIU-FT&0.882&0.858&0.836&0.815&0.824&0.884\\
			\hline
		\end{tabular}
	\end{table}

	\begin{table*}[b!]
		\centering
		\caption{Qualitative results on the validation set. The mistakes are marked in bold.}
		\label{tab:ValSet}
		
		\def\arraystretch{0.95} 
		
		\begin{tabular}{cc}
			\hline
			Categories&Outputs\\
			\hline
			\multirow{9}{*}{Samples from evaluation set}&\multicolumn{1}{|l}{\text{IN: } \textit{we hope that one day it may become a symbol promoting the benefits of ethnic reconciliation}}\\
			\multicolumn{1}{c}{~}&\multicolumn{1}{|l}{\text{CE: we hope that one day can may become a symbol} \textbf{from the fight} \text{of ethnic reconciliation}}\\
			\multicolumn{1}{c}{~}&\multicolumn{1}{|l}{\text{RL: we hope that one day it may become a symbol} \textbf{in the benefits} \text{of ethnic reconciliation}}\\
			\cline{2-2}
			\multicolumn{1}{c}{~}&\multicolumn{1}{|l}{\text{IN: } \textit{and i refer once again to the declarations which we have heard in recent days}}\\
			\multicolumn{1}{c}{~}&\multicolumn{1}{|l}{\text{CE: } \textbf{and that anything cannot again} \text{to the declarations which we have heard in recent days}}\\
			\multicolumn{1}{c}{~}&\multicolumn{1}{|l}{\text{RL: } \text{and i refer once again to the} \textbf{rights} \text{which we have heard in recent days}}\\
			\cline{2-2}
			\multicolumn{1}{c}{~}&\multicolumn{1}{|l}{\text{IN: } \textit{that is why full employment actually needs to be right at the top of the agenda}}\\
			\multicolumn{1}{c}{~}&\multicolumn{1}{|l}{\text{CE: that is why} \textbf{high} \text{employment} \textbf{people} \text{needs to be right at the top of the agenda}}\\
			\multicolumn{1}{c}{~}&\multicolumn{1}{|l}{\text{RL: that is why full employment} \textbf{really} \text{needs to be right at the top of the agenda}}\\
			\cline{1-2}
			\multirow{3}{*}{Failure case}&\multicolumn{1}{|l}{\text{IN: } \textit{this means that the basic principle must be that the price should represent the real cost of the water used}}\\
			\multicolumn{1}{c}{~}&\multicolumn{1}{|l}{\text{CE: this means that the basic principle must be that the price should represent the real cost of the} \textbf{water industry}}\\
			\multicolumn{1}{c}{~}&\multicolumn{1}{|l}{\text{RL: this means that the basic principle must be that the price should represent the real cost of the} \textbf{commission}}\\			
			\hline
		\end{tabular}
	\end{table*}

	Since in the decoding stage both Semantic-JSCC and SemanticRL-SCSIU share the same softmax policy (given in (\ref{sample_decoder}) and (\ref{sample_decoder-SCSIU}) respectively) and the same objective (\ref{def_Gt}), the convergence difference between JSCC and SCSIU is influenced by the TX-side gradient estimation process, that is:
	\begin{subequations}
	\label{SeamnticRL-grad}
	\begin{align}
	&\mathop{\textrm{grad}} \mathcal{H}_{TX} = h^{-1}_{JSCC} \cdot \left( \textrm{grad} \, \mathcal{H}_{RX} \right) \notag  \\
	&\Rightarrow \textrm{grad variance} \\
	&\Theta(\mathcal{F}_{RX} \mathcal{F}_{TX}, \boldsymbol{m}) = h^{-1}_{SCSIU} \cdot \Theta(\mathcal{F}_{RX} \mathcal{H} \mathcal{F}_{TX}, \boldsymbol{m}) \notag \\
	&\Rightarrow \textrm{reward variance}
	\end{align}
	\end{subequations}%
	\noindent
	where ``$\textrm{grad} \ \mathcal{H}_{TX}$'' denotes the gradient in the TX side (just before $\mathcal{H}$), and ``$\textrm{grad} \ \mathcal{H}_{RX}$'' denotes that in the RX side (just after $\mathcal{H}$). $h^{-1}_{(\cdot)}$ is the non-linear inverse channel response function.
	
	From (\ref{SeamnticRL-grad}), we learn that the gradient estimation in JSCC (by simulating with differentiable Gaussian or Fading channel) and reward estimation (by directly execute feed forward propagation) in SCSIU is both \emph{unbiased}, but with certain \emph{variance}. This means that both these approaches can theoretically converge to their optimal parameters, except for some details like convergence speed and error bound, which are also related to estimation variance. It is also revealed that relaxing the differentiability of $\mathcal{H}$ can generally lead to a slower convergence as a differentiable channel significantly simplifies the estimation process. We highlight that through the introduced self-critic rewarding mechanism, SemanticRL offers a direct, precise and low-variance estimation, which facilitates a practical and large-scale development. 
	
	We provide the quantitative results for SemanticRL-SCSIU in Table \ref{tab:SCSIU_res}. As we may notice that this SCSIU variant (labeled as SCSIU-PT as it shares the same pre-trained initial parameter as the JSCC variant) learns slower and probably may not yield a comparable result within our time limit. This is not surprising as JSCC training is \emph{de facto} a significantly simplified scheme, while SCSIU takes longer time to learn from an unknown environment. To better examine the performance upper bound, we have also examined the results derived from a fine-tuning on SemanticRL-JSCC (labeled as SCSIU-FT). Table \ref{tab:SCSIU_res} shows that the SCSIU scheme can eventually produce a comparable result if trained with sufficient time. That said, SemanitcRL-SCSIU provides a practical, large-scale, and proof-of-the-concept step towards a semantics-oriented, wireless, and complete solution for semantic transmission. These important properties, however, are commonly neglected in existing works.

	\subsection{A Closer Look at the Learned Semantic Representation}

	\begin{table*}[t]
		\centering
		\caption{More detailed real-life evaluations with typical semantic ablation.}
		\label{tab:FullEval}
		\def\arraystretch{0.95} 
		\begin{tabular}{cc}
			\hline
			Categories&Outputs\\
			\hline
			
			\multirow{7}{*}{(a) Arbitrary corpus-unrelated input}&\multicolumn{1}{|l}{\textit{This is a typical sentence used to check the performance}}\\
			\multicolumn{1}{c}{~}&\multicolumn{1}{|l}{\text{CE1: this is a substantial important opportunity to check the debates}}\\
			\multicolumn{1}{c}{~}&\multicolumn{1}{|l}{\text{CE2: this is a typical important problem to check the performance}}\\
			\multicolumn{1}{c}{~}&\multicolumn{1}{|l}{\text{CE3: this is a typical important task to check the performance}}\\
			\multicolumn{1}{c}{~}&\multicolumn{1}{|l}{\text{RL1: this is a typical sentence used to check the judge}}\\
			\multicolumn{1}{c}{~}&\multicolumn{1}{|l}{\text{RL2: this is a typical sentence used to check the performance}}\\
			\multicolumn{1}{c}{~}&\multicolumn{1}{|l}{\text{RL3: this is a typical amount used to check the performance}}\\
			\hline

			\multirow{7}{*}{(b) Contextual inference}&\multicolumn{1}{|l}{\textit{This is a typical <UNK> used to check the performance}}\\
			\multicolumn{1}{c}{~}&\multicolumn{1}{|l}{\text{CE1: this is a typical clear task to check the performance}}\\
			\multicolumn{1}{c}{~}&\multicolumn{1}{|l}{\text{CE2: this is a typical common challenge to check the performance}}\\
			\multicolumn{1}{c}{~}&\multicolumn{1}{|l}{\text{CE3: this is a typical legal challenge to check the performance}}\\
			\multicolumn{1}{c}{~}&\multicolumn{1}{|l}{\text{RL1: this is a typical basic problem to check the performance}}\\
			\multicolumn{1}{c}{~}&\multicolumn{1}{|l}{\text{RL2: this is a typical general problem to check the performancee}}\\
			\multicolumn{1}{c}{~}&\multicolumn{1}{|l}{\text{RL3: this is a typical correct approach to liberalisation the performance}}\\
			\hline

			\multirow{7}{*}{(c) Complex task}&\multicolumn{1}{|l}{\textit{This is exactly a long sentence with complex structure which might be a challenge for both}}\\
			\multicolumn{1}{c}{~}&\multicolumn{1}{|l}{\text{CE1: this is exactly a long sentence with complex structure which might be a challenge for both}}\\
			\multicolumn{1}{c}{~}&\multicolumn{1}{|l}{\text{CE2: in it those of others await to humanitarian considerations which might be a challenge for both}}\\
			\multicolumn{1}{c}{~}&\multicolumn{1}{|l}{\text{CE3: this is exactly a long sentence with environmental considerations it would be a challenge for both}}\\
			\multicolumn{1}{c}{~}&\multicolumn{1}{|l}{\text{RL1: this is exactly a long sentence with serious confusion which might be a challenge for both}}\\
			\multicolumn{1}{c}{~}&\multicolumn{1}{|l}{\text{RL2: this is exactly a long sentence with complex relations which might be a challenge for the}}\\
			\multicolumn{1}{c}{~}&\multicolumn{1}{|l}{\text{RL3: this is exactly a long sentence with complex structure which might be a challenge for both}}\\
			\hline
			
			\multirow{7}{*}{(d) Semantic perseverance}&\multicolumn{1}{|l}{\textit{I have just brought a yellow banana}}\\
			\multicolumn{1}{c}{~}&\multicolumn{1}{|l}{\text{CE1: i have just brought a wound in}}\\
			\multicolumn{1}{c}{~}&\multicolumn{1}{|l}{\text{CE2: i have just brought a motorway from}}\\
			\multicolumn{1}{c}{~}&\multicolumn{1}{|l}{\text{CE3: i have just brought a vessel from}}\\
			\multicolumn{1}{c}{~}&\multicolumn{1}{|l}{\text{RL1: i have just brought a european banana}}\\
			\multicolumn{1}{c}{~}&\multicolumn{1}{|l}{\text{RL2: i have just brought a european neighbouring}}\\
			\multicolumn{1}{c}{~}&\multicolumn{1}{|l}{\text{RL3: i have just brought a european banana}}\\
			\hline

			\multirow{7}{*}{(e) Semantic attack}&\multicolumn{1}{|l}{\textit{A man is holding a giant elephant on his hand}}\\
			\multicolumn{1}{c}{~}&\multicolumn{1}{|l}{\text{CE1: a man is holding a bus broke on his hand}}\\
			\multicolumn{1}{c}{~}&\multicolumn{1}{|l}{\text{CE2: a man is holding a giant elephant on his hand}}\\
			\multicolumn{1}{c}{~}&\multicolumn{1}{|l}{\text{CE3: a man is holding a bus unbearable on his hand}}\\
			\multicolumn{1}{c}{~}&\multicolumn{1}{|l}{\text{RL1: a man is holding a giant in on his hand}}\\
			\multicolumn{1}{c}{~}&\multicolumn{1}{|l}{\text{RL2: a man is holding a manner in on his hand}}\\
			\multicolumn{1}{c}{~}&\multicolumn{1}{|l}{\text{RL3: a man is holding a giant this on his hand}}\\
			\hline

		\end{tabular}

	\end{table*}

	In addition to the numerical experiments that provide a holistic view of our proposed SemanticRL, it is critical to understand how the model behaves in real communication scenarios as well. In Table \ref{tab:ValSet}, we provide a number of examples from the validation set. The key mistakes are marked in bold for a better comparison. 
	As to the first three samples for example, we find that the proposed model better expresses the main ideas of a sentence, and is able to choose a semantically similar word even when mistakes do occur. However, the drawback of using CIDEr-D as the reward function (our default setting) is also clear: it encourages the model to pay more attention to the middle part of a sentence (words in the middle share more \ngram weight). As a result, the resulting model lacks certain diversity near where a sentence concludes, as seen from the failure case. We believe that with undergoing research on the semantic similarity, combined with more well-designed training methods, this issue can be handled properly in the near future.

	Another important concern comes from the generalization capability. In this experiment, we consider several typical input sentences and conduct semantic level ablations so as to better understand the learned strategy. For each sentence, we pass it through the channel for three times to diminish the randomness, as shown in Table \ref{tab:FullEval}. In the initial experiment, we feed a simple sentence into both the CE-based model and the proposed SemanticRL. As can be found in the decoded results, SemanticRL not only succeeds in catching the key idea, but also behaves more stably; while the CE baseline generates more unrelated words. Then, we seek to increase the difficulty by masking a key word in this sentence to see if the model can still perform meaningful inference on missed word by looking at the contexts. To further examine the contextual understanding ability, we have also studied a more complex sentence in Table \ref{tab:FullEval}(c). The experimental results all suggest that our model consistently behaves better and captures closer semantic meaning. We also notice that this model may not behave properly when the contextual information is seldom seen in the training set as Table \ref{tab:FullEval}(d) tells, where scene-specific tokens like ``european'' are frequently generated. In extreme situations, similarly, if forced to transmit semantically irrational sentences, SemanticRL may behave poorer than the CE baseline. Although a little disappointing, this phenomenon is indeed consistent with our goal to transmit messages that sound to be ``true'' at the semantic level.

	\subsection{Extension to Other Semantic Communication Tasks}

	Considering a broader case of semantic communication, we claim that by slightly modifying the decision process and learning details, it is convenient to introduce the proposed SemanticRL paradigm on other media (\eg, pictures), so long as one can provide a meaningful semantic measurement. To better demonstrate this viewpoint, we provide an example in this section where we show how to transmit images in a SemanticRL manner.

	By formulating the decoding process as progressively adding or subtracting a small number for each pixel value, we can simply define the intermediate reward as the MSE gain between the previous image and the current one. This leads to a new parametric setting: the intermediate reward function is given in (\ref{reward_image}); and $G^{(t)}$ is then defined in a more general way by setting the discount factor $\gamma=0.99$.
	
	\begin{equation}
		r_i^{(t)} = (I_i-\hat{I_i}^{(t)})^2 - (I_i-\hat{I_i}^{(t+1)})^2  \quad I_i\in I
		\label{reward_image}
	\end{equation}
	where $I_i$ indicates any single pixel in an image. This reward function can be interpreted as to encourage the decisions that lead to a lower MSE. The major difference as compared to the above mentioned sentence-transmission task is to use a globally shared policy for all pixels, and turns the single-agent RL learning into a multi-agent RL learning task, similar as PixelRL \cite{furuta2019pixelrl}.
	
	We illustrate how this image-targeted model learns and behaves in \Figure \ref{Fig:ImageTrans}. The model is trained on MNIST dataset\footnote{\url{http://yann.lecun.com/exdb/mnist/}} with the weight of encoder and the first layer of decoder initialized from a regularly pre-trained one. All the images are quantized every 25.5 pixels for simplicity (so that after normalization, one pixel has 10 values $\{0, 0.1, 0.2, ..., 0.9\}$ to take). We set the initially decoded image $I^{(0)}$ as one filled with pixel value 0.5 and let each action adjust the pixel value by increasing 0.1 or decreasing 0.1, or keeping it unchanged, so that all images could be recovered in five time steps. Note that here we attempt to provide a simple yet proof-of-the-concept example; one may further conduct more complex tasks such as directly optimize the semantic segmentation or detection/reconstruction process with task-specific non-differentiable reward functions like intersection over union (IOU), semantic classification accuracy \etc.

	\begin{figure}[t]
		\centering
		\includegraphics[width=8.8cm]{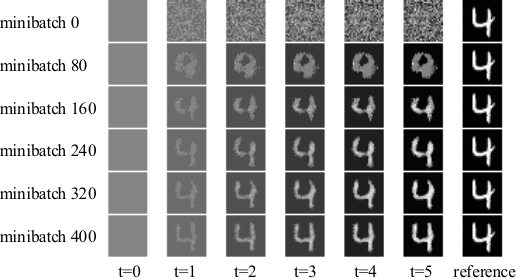}
		\caption{An example on image-transmission extension, where the model is transformed into a semantic agent and gradually learns to recover input semantics by acting progressively.}
		\label{Fig:ImageTrans}
	\end{figure}	
	
	\section{Limitations and Future Work}
	
	Despite the aforementioned advantages, we note that an RL-based learning framework undoubtedly adds more training complexity owing to the possible instability and the time-consuming sequential decoding process. Nevertheless, efforts on JSCC-based end-to-end communication take several years to become a practical system, which we think also applies for this new mechanism. In practice, a training time-performance balance is suggested to be considered in resource-limited communication scenarios, and makes the current scheme more suitable for a centralized, server-level deployment. As another important factor, the dependency on well-defined semantic similarity metrics remains a task-specific and challenging problem. Though in this paper we have provided a practical solution that bridges the semantic gap, we note that more researches on the semantic similarity, and typically information-based interpolations deserve an in-depth survey.

	\section{Conclusion}

	We show in this paper how to develop a universal semantic communication system by optimizing semantic similarity in a reinforcement learning manner. Both the studied semantic objective training and non-differentiable optimization is less or have not been fully investigated in existing works. 
	The proposed solution - SemanticRL allows any semantic similarity metric as the reward function and explores the first concrete systematic shift from reliable communication to semantic level transmission. It better captures the semantic essence of transmitted messages, and is shown to be more robust for unknown noisy environments. The decoupled SCSIU variant, on the other hand provides a complete solution for large-scale wireless semantic transmission and might have independent interest to the community. Meanwhile, the limitations and possible generalization examples are also exemplified for a future study. To the best of our knowledge, researches on semantic similarity, particularly a reinforcement learning-based practical system in semantic communication, are rather limited. We hope our work could shed some new light on these problems. 

	\input{Appendix}

	\bibliographystyle{IEEEtran}
	\bibliography{SemanticRLv2}

	
	
	
	
	

\end{document}

%% file: Tab_Notation.tex
\begin{table}[t]
	\centering
	\caption{Notations used in this paper.}
	\label{tab:notations}
	
	\def\arraystretch{1.0} 
	
	\begin{tabular}{ll}			
		\hline
		Notation & Definition \\
		\hline
		$\boldsymbol{m}, \bh{m}$&Input message and decoded message \\
		$\Theta$ & Semantic similarity metric \\
		$\phi, \theta$ & Parameters for encoder and decoder respectively\\
		$\mathcal{F}_{TX}, \mathcal{F}_{RX}$ & Abstract encoder and decoder \\
		$\mathcal{H}$ & Random channel \\
		$h^{-1}$ & Inverse channel response function \\
		$(\widetilde{~~})$ & Sampling \\
		$T, N$ & The length of input and output messages repetitively \\
		$w^{(t)}, \hat{w}^{(t)}$ & The $t$-th word in the input and decoded message \\ 
		$\mathcal{W}$ & The dictionary \\
		$V$ & Dimension of the dictionary \\
		$d$ & Dimension of message embedding \\

		$g(\cdot)$ & Term frequency inverse document frequency weight \\
		$\mathcal{N}(\boldsymbol{\mu}, \boldsymbol{\Sigma})$ & Gaussian distribution \\
		$\mathcal{MN}([p_1, p_2 ...])$ & Multinomial distribution \\			
		$j$ & The imaginary unit \\
		
		$p(\cdot)$ & Probability of choosing a single word \\
		$P(\cdot)$ & Probability of taking a complete trajectory \\
		$M$ & Number of parallel samples \\
		$r_t$ & Reward at time $t$ \\ 
		$s_t$ & State at time $t$ \\
		$a_t$ & Action at time $t$ \\
		$G_t$ & The return at time $t$ \\
		$\mathcal{S}, \mathcal{A}$ & The state space and action space \\
		$\gamma$ & Discount factor \\
		$\sigma$ & Scale factor that controls the exploration process \\
		$\pi$ & Policy \\

		$\alpha$ & Learning rate\\
		$J$& The objective function \\
		\hline
	\end{tabular}
\end{table}

%% file: Alog_SemanticRL-JSCC.tex
\begin{figure}[!t]
	\renewcommand{\algorithmiccomment}{\textbf{// }}
	\renewcommand{\algorithmicrequire}{\textbf{Input:}}
	\renewcommand{\algorithmicensure}{\textbf{Output:}}
	\removelatexerror
	\begin{algorithm}[H]
		\caption{SemanitcRL-JSCC (joint source-channel coding)}
		\begin{algorithmic}[1]
			\REQUIRE Batch size 64, initial learning rate $\alpha=1e-3$, self-critic samples $M=5$, input sequence $\boldsymbol{m}$, pre-training epoch $E_p=87$, end epoch $E_e=200$, semantic similarity metric $\Theta$
			\ENSURE Encoder parameter $\phi$, decoder parameter $\theta$
			
			\COMMENT{\textbf{Pre-training stage}}
			\FOR{epoch=1:$E_p$}
			\STATE Sample a batch of data, update $\alpha$ as described in Section \ref{Training}
			\STATE TX encodes $\boldsymbol{m}$ into its embedding $\mathcal{F}_{\textrm{TX}}(\boldsymbol{m})$. Do average power normalization, and send through a noisy channel 
			\STATE RX decodes, and calculates loss $L_{CE}$ using (\ref{CE_loss})
			\STATE Update $\phi$ and $\theta$ jointly with $\phi; \theta \gets \phi; \theta -\alpha\cdot \nabla L_{CE}$
			\ENDFOR
			
			\COMMENT{\textbf{RL-based JSCC training stage}}
			\FOR{epoch=$E_p$+1:$E_e$}
			\STATE Sample a batch of data, update $\alpha$ as described in Section \ref{Training}
			\STATE TX encodes $\boldsymbol{m}$ into its embedding $\mathcal{F}_{\textrm{TX}}(\boldsymbol{m})$. Do average power normalization, and send through a noisy channel
			\FOR{i=1:M}
			\STATE Sample one trajectory as in (\ref{sample_decoder})
			\ENDFOR
			\STATE Calculate the semantic similarity score $\Theta$ for each simulation
			\STATE Calculates the gradient of $J(\phi; \theta)$ using (\ref{self-critic})
			\STATE Update $\phi$ and $\theta$ jointly with $\phi; \theta \gets \phi; \theta + \alpha\cdot \nabla J(\phi; \theta)$
			\ENDFOR
			
			\COMMENT{\textbf{Finish training}}
			\STATE Return the parameter $\langle \phi, \theta \rangle$
		\end{algorithmic}
	\end{algorithm}
	\vspace{-4mm}
\end{figure}

%% file: Alog_SemanticRL-SCSIU.tex
\begin{figure}[!t]
	\renewcommand{\algorithmiccomment}{\textbf{// }}
	\renewcommand{\algorithmicrequire}{\textbf{Input:}}
	\renewcommand{\algorithmicensure}{\textbf{Output:}}
	\begin{algorithm}[H]
		\caption{SemanitcRL-SCSIU (self-critic stochastic iterative updating)}
		\begin{algorithmic}[1]
			\REQUIRE Batch size 64, initial learning rate $\alpha=1e-4$, self-critic samples $M=5$, input sequence $\boldsymbol{m}$, pre-training epoch $E_p=87$, end epoch $E_e=300$, semantic similarity metric $\Theta$, update flag = 0, scale factor $\sigma=0.1$
			\ENSURE Encoder parameter $\phi$, decoder parameter $\theta$
			
			\COMMENT{\textbf{Pre-training stage}}
			\STATE This stage keeps the same as SemanticRL-JSCC
			
			\COMMENT{\textbf{RL-based SCSIU training stage}}
			\FOR{epoch=$E_p$+1:$E_e$} 
			\STATE Sample a batch of data
			
			\IF{flag=0}
				\STATE 	\COMMENT{\textbf{Training Encoder}}
				\STATE Freeze RX (and also use argmax decoding so that the decoded results are deterministic). TX encodes $\boldsymbol{m}$ into its embedding.
				\STATE Do average power normalization
				\FOR{i=1:M}
					\STATE Sample one message embedding according to (\ref{sample_encoder}), do power normalization, and send through a noisy channel
				\ENDFOR
				\STATE RX decodes with argmax strategy and calculates the semantic scores of these samples
				\STATE (Optional) accumulate gradient to reduce the variance
				\STATE Update $\phi$ with Gaussian policy as in (\ref{SeamnticRL-SCSIU}a) and (\ref{Grad-SCSIU}a)
				
			\ELSE
				\STATE \COMMENT{\textbf{Training Decoder}}	
				\STATE Freeze TX (and also set $\boldsymbol \Sigma=\boldsymbol{0}$ so that the embeddings are deterministic); do average power normalization and execute feed forward through the channel
				\FOR{i=1:M}
					\STATE RX samples one trajectory according to (\ref{sample_decoder-SCSIU}) and calculates the semantic scores of these samples
				\ENDFOR
				\STATE (Optional) accumulate gradient to reduce the variance
				\STATE Update $\theta$ with softmax policy as in (\ref{SeamnticRL-SCSIU}b)  and (\ref{Grad-SCSIU}b)
			\ENDIF
			\STATE Flip the update flag
			
			\ENDFOR
			
			\COMMENT{\textbf{Finish training}}
			\STATE Return the parameter $\langle \phi, \theta \rangle$
		\end{algorithmic}
	\end{algorithm}
\end{figure}

%% file: Appendix.tex
\appendices
\section{Proof of Theorem 1}

	Denoted by $P(\bh{m}| \phi; \theta)$ the probability of taking a complete trajectory $\bh{m}$, we first  expand the expectation in (\ref{objectivef}): 
		
	\begin{equation}
		\begin{aligned}
			J(\phi; \theta) &= \mathop{\mathbb{E}}_{\hat{w}^{(1)},...,\hat{w}^{(N)}} \left[ \sum_{t=1}^{N}  r^{(t)} \right] \\
			&= \sum_{\bh{m}} P(\bh{m}| \phi; \theta) \left( \sum_{t=1}^{N} r^{(t)} \right) \\
		\end{aligned}
		\label{APP_expansion}
	\end{equation}
	
	Calculating the gradient on both sides, we have:
	
	\begin{equation}
		\begin{aligned}
			\nabla J(\phi; \theta) &= \nabla_{\phi,\theta} \left[ \sum_{\bh{m}} P(\bh{m}| \phi; \theta) \left( \sum_{t=1}^{N} r^{(t)} \right) \right] \\
			&= \sum_{\bh{m}} \left[ \nabla_{\phi, \theta} P(\bh{m}| \phi; \theta) \left( \sum_{t=1}^{N} r^{(t)} \right) \right] \\
			&= \sum_{\bh{m}} \left[ P(\bh{m}| \phi; \theta) \nabla_{\phi, \theta} \log\left(P(\bh{m}| \phi; \theta)\right) \left( \sum_{t=1}^{N}  r^{(t)} \right) \right] \\
			&= \mathop{\mathbb{E}}_{\bh{m}} \left[ \nabla_{\phi, \theta} \log\left(P(\bh{m}| \phi; \theta)\right) \left( \sum_{t=1}^{N}  r^{(t)} \right) \right]
		\end{aligned}
		\label{APP_derivative_formu}
	\end{equation}
	where the second equation holds by 1) the fact that \textit{return} does not depend on $\theta$ (see Appendix A1 in \cite{williams1992simple}), and 2) swapping the summation and derivation will not change the value here; the third equality leverages the log-trick that $\nabla \log\left(\boldsymbol{x}\right) = \nabla \boldsymbol{x} \, / \boldsymbol{x}$.

	In practice, we can use a one-time Monte-Carlo rollout for the estimation of the expectation on $\bh{m}$, that is, 
	
	\begin{equation}
		\nabla J(\phi; \theta) \approx \nabla_{\phi, \theta} \log\left(P(\bh{m}| \phi; \theta)\right) \left( \sum_{t=1}^{N} r^{(t)} \right)
		\label{APP_MC-derivative}
	\end{equation}
	
	Further, considering the fact that one complete trajectory is composed of $N+1$ consecutive actions (including an ending token), the probability of generating a sequence $P(\bh{m};\theta)$ can further be expanded as \mbox{$P(\bh{m};\theta)=p(s^{(0)}) \prod_{t=1}^{N}\left[ \pi_{\phi, \theta}(\hat{w}^{(t)}|s^{(t)}) p(s^{(t)}|s^{(t-1)},a^{(t-1)}) \right]$}. Then, for (\ref{APP_MC-derivative}) we write
	
	\begin{equation}
		\begin{aligned}
			\nabla J(\phi; \theta) & \approx  \nabla_{\phi, \theta} \log\left( p(s_0)
			\prod_{t=1}^{N}  \pi_{\phi, \theta}(\hat{w}^{(t)}|s^{(t)})  \right) \left( \sum_{t=1}^{N} r^{(t)} \right) \\ 
			&= \sum_{t=1}^{N} \nabla_{\phi, \theta} \log \pi_{\phi, \theta}(\hat{w}^{(t)}|s^{(t)})  \left( \sum_{t=1}^{N}  r^{(t)} \right)\\
		\end{aligned}
		\label{APP_probab_traj}
	\end{equation}

	Recall that the reward $r^{(t)}$ here is sparse as given by (\ref{reward}), we then substitute $\sum_{t=1}^{N} r^{(t)}$ with $\Theta(\boldsymbol{m},\bh{m})$, and that ends the proof with:
	
	\begin{equation}
		\nabla J(\phi; \theta) \approx \sum_{t=1}^{N} \nabla_{\phi, \theta} \log \pi_{\phi, \theta}(\hat{w}^{(t)}|s^{(t)}) \Theta(\boldsymbol{m},\bh{m})  
		\label{APP_probab_traj_subs}
	\end{equation}

\hfill $\blacksquare$

\section{Proof of Theorem 2}

\noindent
1) Derivation of (\ref{Grad-SCSIU}a)

For ease of notation, we represent RX's output at time step $t$ as $\boldsymbol{z}\in \mathbb{R}^V$ where $V$ is the dictionary size. A softmax layer is then exposed to $\boldsymbol{z}$:
\begin{equation}
	a_k = \frac{e^{z_k}}{\sum_{i=1}^{V} e^{z_i}}
	\label{APP_B_softmax}
\end{equation}
so that $\boldsymbol{a} = \left[p(\hat{w}_1^{(t)}), p(\hat{w}_2^{(t)})...p(\hat{w}_V^{(t)})\right]^T$ is bounded in $[0,1]$, corresponding to our notations in (\ref{sample_decoder}).

Then we denote $L = \log(\pi_{\theta}(\hat{w}^{(t)}|s^{(t)}))$. According to the chain rule, $\frac{\partial L}{\partial z_k} = \frac{\partial L}{\partial a_j} \cdot \frac{\partial a_j}{\partial z_k}$. 

For the first term in RHS, we have $\frac{\partial L}{\partial a_j} = \frac{1}{a_j}$. For the second term, we have:

\begin{equation}
\begin{aligned}
\frac{\partial a_j}{\partial z_k} &= e^{z_j} \cdot \frac{-e^{z_k}}{(\sum_{i=1}^{V} e^{z_i})^2}  \quad \textrm{if} \quad i \neq j \\
&= -a_j \cdot a_k\\
\frac{\partial a_j}{\partial z_k} &= \frac{e^{z_j}\cdot(\sum_{i=1}^{V} e^{z_i}) - e^{z_j}\cdot e^{z_j}}{(\sum_{i=1}^{V} e^{z_i})^2}  \quad \textrm{if} \quad i = j  \\
&= a_j - (a_j)^2
\end{aligned}
\end{equation}

Combining these two terms, we get:
\begin{equation}
\begin{aligned}
\frac{\partial L}{\partial z_k} &=
\begin{cases}
-a_k  \quad \textrm{if} \quad i \neq j  \\
1-a_k  \quad \textrm{if} \quad i = j  \\
\end{cases}
\\
&= \mathbbm{1} - \left[p(\hat{w}_1^{(t)}), p(\hat{w}_2^{(t)})...p(\hat{w}_V^{(t)})\right] \\
\end{aligned}
\end{equation}

\hfill $\blacksquare$

\noindent
2) Derivation of (\ref{Grad-SCSIU}b)

A Gaussian distribution with mean value $\boldsymbol{\mu}^{d\times 1}$ and covariance matrix $\boldsymbol{\Sigma}^{d\times d}$ has the following probability density function:
\begin{equation}
\pi(\boldsymbol{x};\boldsymbol{\mu}, \boldsymbol{\Sigma}) = \frac{1}{(2\pi)^{d/2} \sqrt{\det{\boldsymbol{\Sigma}}}} \exp \left(-\frac{1}{2}(\boldsymbol{x}-\boldsymbol{\mu})^T \boldsymbol{\Sigma}^{-1} (\boldsymbol{x}-\boldsymbol{\mu})\right)
\end{equation}

The log-likelihood function is given by

\begin{equation}
\begin{aligned}
&\log \pi(\boldsymbol{x};\boldsymbol{\mu}, \boldsymbol{\Sigma}) =  \\
&-\frac{d}{2}\log{2\pi} -\frac{1}{2}\log(\det{\boldsymbol{\Sigma}}) - \frac{1}{2} (\boldsymbol{x}-\boldsymbol{\mu})^T \boldsymbol{\Sigma}^{-1} (\boldsymbol{x}-\boldsymbol{\mu})
\end{aligned}
\end{equation}

Then the gradient can be calculated as

\begin{equation}
	\nabla_{\phi} \log \pi(\boldsymbol{x};\boldsymbol{\mu}, \boldsymbol{\Sigma}) = \left[\boldsymbol{x}-\boldsymbol{\mu}\right]^T \boldsymbol{\Sigma}^{-1}   \left[\nabla_{\phi}\boldsymbol{\mu}\right]
\end{equation}

When $\boldsymbol{x}=\widetilde{\mathcal{F}_{TX}}(\boldsymbol{m})$, and $\boldsymbol{\mu}=\mathcal{F}_{TX}(\boldsymbol{m})$, we get the equation in (\ref{Grad-SCSIU}b). %

\hfill $\blacksquare$
